\documentclass[10pt,twocolumn,letterpaper]{article}

\usepackage{wacv}
\usepackage{times}
\usepackage{epsfig}
\usepackage{graphicx}
\usepackage{amsmath}
\usepackage{amssymb}

\usepackage{color}
\usepackage{booktabs}
\usepackage{multirow}
\usepackage{amsfonts}
\usepackage{bm}
\usepackage[rightcaption]{sidecap}

\definecolor{newcolor}{rgb}{.8,.349,.1}
\definecolor{gray1}{rgb}{0.84,0.84,0.84}
\definecolor{gray2}{rgb}{1,0.89,0.75}
\definecolor{clr3}{rgb}{0.95,0.95,0.95}
\definecolor{clr4}{rgb}{0.96,0.96,0.86}

\def\wacvPaperID{0595} 

\wacvfinalcopy 

\ifwacvfinal
\def\assignedStartPage{0001} 
\fi

\ifwacvfinal
\usepackage[breaklinks=true,bookmarks=false]{hyperref}
\else
\usepackage[pagebackref=true,breaklinks=true,colorlinks,bookmarks=false]{hyperref}
\fi

\ifwacvfinal
\else
\fi

\begin{document}

\title{Attention Guided Semantic Relationship Parsing for Visual Question Answering}

\author{Moshiur Farazi$^{1,2}$\thanks{Corresponding author}, \quad Salman Khan$^{3,1}$, \quad Nick Barnes$^{1}$\\
$^{1}$RSEEME, Australian National University, \quad $^{2}$Data61--CSIRO, Canberra, Australia\\
$^{3}$Mohamed bin Zayed University of Artificial Intelligence, Abu Dhabi, UAE\\
{\tt\small firstname.lastname@anu.edu.au}
}


\maketitle

\begin{abstract}
Humans explain inter-object relationships with semantic labels that demonstrate a high-level understanding required to perform complex Vision-Language tasks such as Visual Question Answering (VQA).
However, existing VQA models represent relationships as a combination of object-level visual features which constrain a model to express  interactions between objects in a single domain, while the model is trying to solve a multi-modal task.
In this paper, we propose a general purpose semantic relationship parser which generates a semantic feature vector for each subject-predicate-object triplet in an image, and a Mutual and Self Attention (MSA) mechanism that learns to identify  relationship triplets that are important to answer the given question. 
To motivate the significance of semantic relationships, we show an oracle setting with ground-truth relationship triplets, where our model achieves a $\sim$25\% accuracy gain over the closest state-of-the-art model on the challenging GQA dataset. Further, with our semantic parser, we show that our model outperforms other comparable approaches on VQA and GQA datasets.
\end{abstract}

\section{Introduction}
Humans can perform high-level reasoning over an image by seamlessly identifying the objects of interest and associated relationships between them. Although objects are central to scene interpretation, they cannot be independently used to develop a holistic understanding of the visual content without considering their mutual relationships. The multi-modal reasoning task of Visual Question Answering (VQA) requires learning precisely encoded relationships between objects. Given the complexity of the task, we advocate for relationship modeling in the semantic space so that a given question can be directly related with the objects and relationships present in an image. Our choice is motivated by two observations. \emph{First}, visual representations for different instances of the same semantic relationship can be very different, making it challenging for the VQA model to relate them with the asked question. \emph{Secondly}, different semantic relationship interpretations can exist for a single visual representation, thereby requiring an enriched mechanism to encode a diverse set of semantic relationships. If a relationship parser can automatically derive representations in semantic space and attend to relevant relations, the above challenges can be simplified for VQA.

\begin{figure*}
\centering
\includegraphics[width=.8\linewidth]{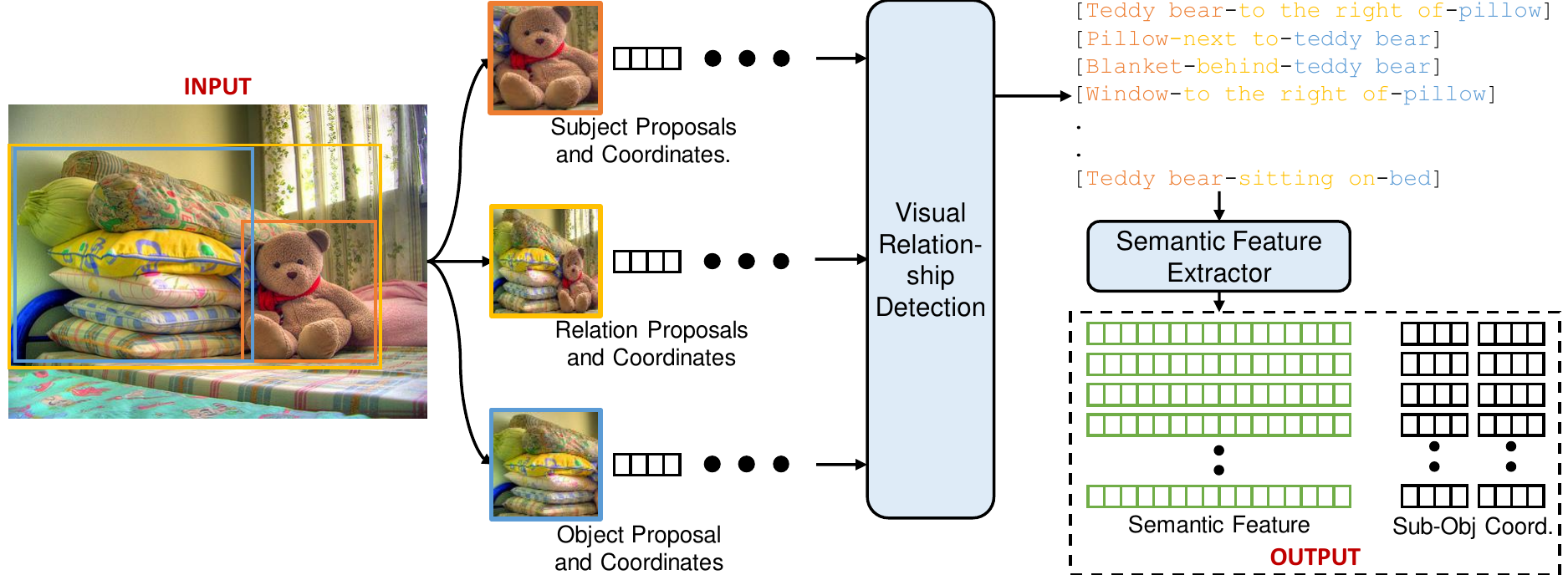}
\caption{{\emph{Our proposed Semantic Relationship Parser (SRP).} Given an image, the SRP module generates relationship triplets through a relationship detector, which are then passed through a semantic feature extractor. Each semantic relationship feature is paired with subject and object box coordinates for visual grounding. Our approach is built upon semantic description of relationships, as opposed to a visual representation in existing works, which allows us to accurately model complex relationships.}}
\label{fig:srp}
\end{figure*}

Based one the hypothesis that a better scene understanding requires a model to generate more discriminative visual and semantic feature representation, recent VQA models employ state-of-the-art visual \cite{he2016deep, hu2018squeeze, ren2015faster} and semantic \cite{pennington2014glove, mikolov2013distributed, devlin2018bert} feature extractors. Specifically, VQA models use information at grid-level \cite{antol2015vqa, benyounescadene2017mutan, yu2017multi}, object-level \cite{Anderson2017up-down, ben2019block, yu2019deep, gao2019dynamic,farazi2020known} or a combination of both \cite{nam2016dual, Farazi_2018_BMVC, farazi_2020_ICPR} to extract visual features in an image without considering the relationships between them. Some recent models address this problem by identifying the most relevant object pairs by learning an attention distribution over them with respect to the question \cite{cadene2019murel, Li_2019_ICCV, Hu_2019_ICCV}. This kind of relationship-aware models achieve better performance compared to the ones that do not consider any kind of relationship. However, as seen in the example shown in Fig.~\ref{fig:srp}, the visual feature representation of \texttt{\small teddy bear} and \texttt{\small pillow} remains the same even though the relationship between them can be different (\eg, \texttt{\small on the right}, \texttt{\small near to}). For higher level reasoning, a visual-semantic model needs to identify these subtle differences, which can only be achieved if the model considers semantic relationship features. 

While attempting to combine semantic relationship features, a VQA model faces three major challenges. \emph{First}, the lack of a visual relationship detector that not only detects arbitrary relationships between object pairs in an image, but also generates semantic features for the detected subject-predicate-object triplets for a downstream task. \emph{Second}, the effectiveness of semantic relationship features over the visual relationship feature in a complex Vision-Language (VL) task such as VQA is not investigated before. \emph{Third}, an effective attention mechanism is required to combine rich features encoding visual, semantic and relationship information to predict the correct answer. In this paper, we contribute towards bridging the gap by addressing these three main challenges. The main contributions of this paper are:
\begin{itemize}\setlength{\itemsep}{-3pt}
\item We propose a general purpose semantic relationship parser that can be used for a complex multi-modal Vision-Language (VL) downstream tasks such as Visual Question Answering.
\item We showcase the effectiveness of using semantic relationship features by reporting superior performance over models employing similar visual relationship features. Further, in an oracle setting where ground-truth relationship labels are available, we obtain a 25\% accuracy gain compared to a SOTA model that only uses visual features.
\item We further propose a Mutual and Self Attention (MSA) mechanism that utilizes both mono-modal self-attention and multi-modal mutual-attention using visual features (from the image) and semantic features (from both question and relationships), and report superior accuracy on the VQAv2 and GQA datasets.
\end{itemize}



\section{Related Works}
\textbf{Vision-Language (VL) pre-training:} Pre-training a model on a different VQA dataset or even a different VL task, to learn a generic visual-linguistic representation by augmented training, has been shown to improve the accuracy VL downstream tasks \cite{su2019vl, lu2016hierarchical, chen_uniter_ECCV2020, zhou2020unified, tan2019lxmert, li2020oscar}.
VL-BERT\cite{su2019vl} and ViLBERT \cite{lu2019vilbert} used transformer models to jointly embed visual and linguistic features and used pretrained models for VL downstream tasks. 
UNITER\cite{chen_uniter_ECCV2020} and Zhou \etal \cite{zhou2020unified} applied an unified approach to learn the pre-trained representation by jointly training a transformer-like architecture. 
LXMERT \cite{tan2019lxmert} and OSCAR\cite{li2020oscar} trained a model to learning a cross-modality representation between more expressive visual and linguistic features on different VL tasks, and fine-tuned the learned representation for downstream VL tasks.
Our approach is orthogonal to such VL pre-training, where we model visual relationships in the semantic domain from subject-relationship-object triplets to leverage the effectiveness modeling visual relationship in semantic domain; and our approach can be easily be included in the VL pre-training schedule.

\textbf{Visual relationships in VQA:} The two major obstacles in utilizing visual relationships in a VQA model are the lack of ground-truth relationship labels and a  way to represent the relationship features.  
Several recent VQA models \cite{xu2017scene, Li_2019_ICCV, Hu_2019_ICCV, zhang2019empirical} resorted to graph neural network approach where the objects pairs represented the nodes and relationship features were represented by some combination of the object features. This approach has two practical limitations, first, it relies heavily on the graph representation and the model's ability to reason over the graph representation. Second, the lack of a real-world VQA dataset that has ground-truth graph representations of images to train and test the models. A few models \cite{teney2016graph, santoro2017simple} tried to capture the relationships from rendered synthetic VQA datasets (Abstract Scene VQAv1 \cite{antol2015vqa}, CLEVR \cite{johnson2016clevr}), which does not generalize well to real scenes. Even though, the Visual Genome \cite{krishna2016visual} dataset has scene graph annotations, the lack of scene graph representations in benchmark VQA datasets (\eg, VQAv1\cite{antol2015vqa}, VQAv2\cite{Goyal_2017_CVPR}, VQA-CP\cite{Agrawal_2018_CVPR}) limits a model's ability to generate graph representations. We adopt to a tangential approach, where we treat the visual relationship feature not as a combination of visual features or a graph, rather as a semantic mono-modal feature representation from its subject, predicate and object labels.

\textbf{Attention models in VQA:} A large portion of the VQA literature focuses on learning a multi-modal representation of image and question features to generate an attention distribution over the input visual feature representation \cite{fukui2016multimodal, vinyals2015show, benyounescadene2017mutan, ben2019block, Farazi_2018_BMVC, yu2017multi, kim2018bilinear}. These approaches have been very successful in learning the multi-modal interactions, however they do not learn mono-modal attention distributions over the inputs themselves \eg, identifying correlation between different image regions or relationships between different words of the question. Inspired by success of self-attention mechanism \cite{vaswani2017attention} in capturing long range dependencies, Yu \etal \cite{yu2019deep} proposed to use self-attention to capture the mono-modal interaction in a VQA setting. However, for achieving high-level visual understanding, one needs to learn both mono-modal and multi-modal interactions, which we propose in this work.


\section{Methods}
Given an image ${I}$ and a natural language question ${Q}$, the task of a VQA model is to predict the answer $\hat{a}$. Let $\bm{v}$ and $\bm{r}$ be the collection of all visual features and semantic relationship features extracted from the image $I$, and $\bm{q}$ be the semantic feature representation of the question $Q$. The VQA problem is typically formulated as a multi-class classification problem: 
\begin{align}
    \hat{a} = \operatorname*{arg\,max}_{a \in \mathcal{A}}  p(a|\bm{v},\bm{q},\bm{r};\bm{\theta}),
    \label{eq:vqa_argmax}
\end{align}
where $\bm{\theta}$ denotes the parameters of the model and $\mathcal{A}$ is a dictionary of candidate answers.

\subsection{Question and Image Feature Extraction}
\label{sec:img_vis_feat_extract}
The traditional approach \cite{fukui2016multimodal, benyounescadene2017mutan, ben2019block, teney2017tips, yu2019deep} for extracting question features for the VQA task is by sourcing pretrained semantic embedding vectors for each question words, concatenating and passing them through a recurrent neural network. The hidden state of the last recurrent block is extracted as the question feature. In contrast, we consider the question as a whole instead of separate word entities, thereby providing better contextual modelling. We use Bidirectional Encoder Representations from Transformers (BERT) \cite{devlin2018bert} where we first tokenize each word of the question and then feed the tokenized question into a Transformer model pretrained for language modeling task. The question feature $\bm{q} \in \mathbb{R}^{m \times d_q}$ is extracted from the last hidden layer of the BERT model, where $m$ is the number of tokens identified in the question and $d_q$ denotes the feature dimension.

We represent the visual features of an input image as a set of  bounding box coordinates and corresponding object-specific features. First, the object proposals are generated using a bottom-up \cite{Anderson2017up-down} attention approach where a pretrained Faster-RCNN~\cite{ren2015faster} model is employed the get the region proposals and extract visual features using a ResNet~\cite{he2016deep} backbone. Following \cite{Anderson2017up-down}, we use an adaptive threshold to select a range of region proposals $l \in [10,100]$ for each image. Further, to visually ground each region proposal we concatenate each region proposal with its bounding box coordinates. Thus, the visual feature representation $\bm{v}\in \mathbb{R}^{l \times (d_v + 4)}$ of image $I$ consists of features of its object proposals $\{\bm{f}_j \in \mathbb{R}^{d_v}\}_{j=1}^l$ and corresponding bounding box coordinates $\{\bm{b}_j \in \mathbb{R}^4\}_{j=1}^l$, where $d_v$ is the object feature dimension.

\subsection{Semantic Relationship Parsing}
\label{sec:srp}
The Semantic Relationship Parser (SRP) module is illustrated in Fig.~\ref{fig:srp}. It has three major components. The \textbf{first} component is  a region proposal network that operates in a similar manner as explained above in Sec.~\ref{sec:img_vis_feat_extract} for visual feature generation. Based on these object-wise features and box coordinates, a visual relationship detector is used in the \textbf{second} stage. 
The visual relationship detector generates subject, relationship and object\footnote{Here, the object refers to the grammatical component of a sentence.} proposals from the region features which in-turn are used to generate a \emph{semantic relationship triplet} set $\mathcal{T} = \{\bm{t}_k\}_{k=1}^q$. 
Each relationship triplet $\bm{t}_k$  consists of class labels predicted for  subject, relationship, and object. 
In order to generate a triplet, a set of candidate subject, relationship and object visual features, denoted by $\bm{f}_s$, $\bm{f}_r$ and $\bm{f}_o$ respectively, are passed through the visual relationship detector. We follow the framework proposed by \cite{zhang2019large}, where we assume a relationship exists only if a subject-object pair exists, not vice versa. Thus the relationship detector learns two mapping functions from visual feature space to semantic space, one for subject/object  and the other one for relationship embedding.

The relationship feature embedding is generated by passing the concatenated version of three visual features $\bm{f}_s$, $\bm{f}_r$ and $\bm{f}_o$ through a two-layer Multi-layer Perceptron (MLP) network. 
The subject and object feature embeddings are generated in parallel by passing them through the same MLP network. On the other hand, class labels of subject, object and relationship are first converted to word vectors and then to semantic feature embeddings by passing them through a small MLP network. Three triplet losses are minimized \cite{zhang2019empirical} to match visual and semantic embedding for subject, object and relationship respectively. 
During inference, word vectors of all subject/object and relationship class labels are passed and a nearest neighbour search is performed to find the desired relationship labels. In practice, we perform the visual relationship detection on the input image as a pre-processing step where we train the visual relationship detector end-to-end on visual relationship dataset (\ie, Visual Genome \cite{krishna2016visual}, VRD~\cite{lu2016visual}) and then run inference on the input images to generate relationship prediction, which consists of subject, object and relationship probability.


\begin{figure*}
\begin{center}
\includegraphics[width=.8\linewidth]{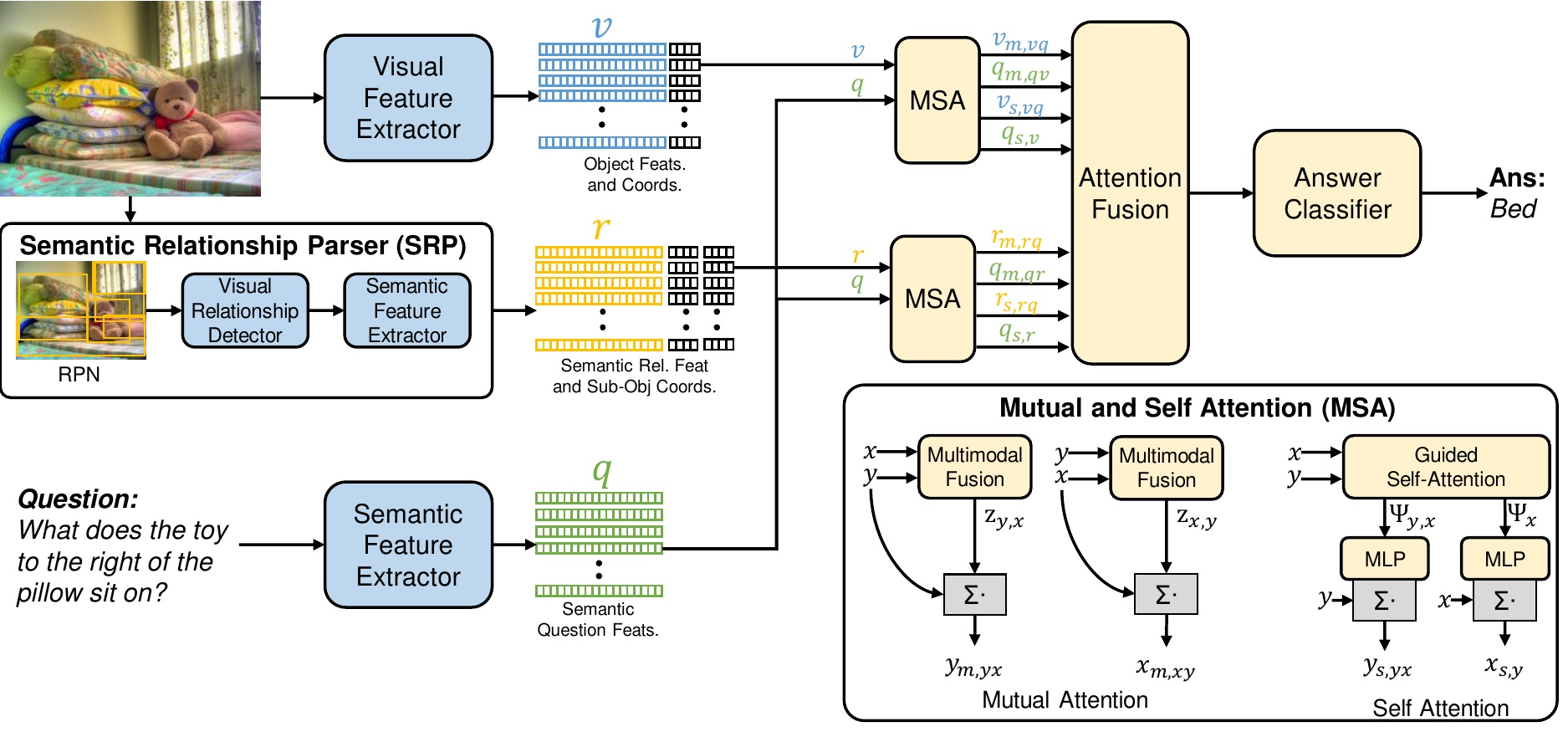}
\end{center}
\caption{\small{\emph{Our proposed Mutual and Self-Attention (MSA) VQA model built on the Semantic Relationship Parser (SRP).} 
The question feature $\bm{q}$ is used alongside visual $\bm{v}$ and relationship  $\bm{r}$ features to generate corresponding mutual and self-attended representations. These rich feature representations are then projected into a common embedding space and sum-pooled to predict the correct answer.
}}
\label{fig:overall_pipeline}
\end{figure*}

The \textbf{third} component of the SRP module is the semantic feature extractor which takes the \emph{filtered} semantic relationship triplets $\mathcal{R}$, subject bounding box $\bm{b}_s$ and object bounding box $\bm{b}_o$ coordinates, and generates visually grounded semantic relationship features as follows,
\begin{align}
    \bm{r} = \mathcal{F}_{\text{BERT}} (\mathcal{R}, \bm{b}_s, \bm{b}_o).
\end{align}
 Here, $\mathcal{F}_{\text{BERT}} $ denotes the BERT model (similar to the one described in Sec.~\ref{sec:img_vis_feat_extract}) for extracting semantic features from the triplets. First, this function takes all entries from the set $\mathcal{R}$ and add{s} a period (`.') after each element. Each relation triplet $\bm{t}_k \in \mathcal{R}$ is now considered a complete sentence which is passed through the BERT model separately alongside its subject and object proposal coordinates. This generates corresponding semantic relationship features $\bm{r} \in \mathbb{R}^{n \times d_r}$ from image $I$, where $d_r$ is the dimension of the hidden feature of the BERT model. 
 
 Notably, the set $\mathcal{R}$ only contains refined relationship triplets obtained after a two stage thresholding and filtering process. At the first stage, we filter out the relationship predictions where the probability of the product of subject and object proposals are higher than a threshold $\alpha$. This ensures that we only select the high-confidence relationships between subject-object pairs. In the next stage, from the remaining relationship predictions, we select only those whose relation probability is higher than $\beta$. Intuitively, we set $\alpha$ at a higher value compared to $\beta$ to ensure that we first get the subject and object instances right, and ask the model to predict various relationships between them. The values of $\alpha$ and $\beta$ are set empirically with the objective to select at least three relationship triplet per image. 
 We further filter out any duplicate relationships and end up with `$n$' relationship predictions per image encoded in refined semantic relationship set $\mathcal{R}$. 



\subsection{Mutual and Self Attention}
\label{sec:msa}
The Mutual and Self Attention (MSA) module consists of two major components. The first component focuses on mutual attention where two separate multimodal fusion operations are performed to learn attention distribution over the input feature vectors. The second component applies self attention on the pair of input features and generates attention distribution over the input features themselves. We illustrate the MSA module in Fig.~\ref{fig:overall_pipeline}. For simplicity lets assume the input to the MSA module are a two feature embeddings $\bm{x} \in \mathbb{R}^{a \times d_x} $ and $\bm{y} \in \mathbb{R}^{b \times d_y}$, which will undergo mutual and self attention. 

\textbf{Mutual Attention:} To capture the complex interaction between $\bm{x}$ and $\bm{y}$ we jointly embed these features by learning a multimodal embedding function. This is achieved by first concatenating the input features and then passing them through a 3 layer MLP network. The MLP model learns to capture the mutual interactions between the input feature vectors and produces a joint feature embedding. For input combinations $(\bm{y},\bm{x})$ and $(\bm{x},\bm{y})$, we have:
\begin{align}
\label{eq:mutual_mlp}
    & \bm{z}_{y,x} = \text{MLP} \, \, [\bm{y} \, \, \oplus \, \, \bm{x}] \in \mathbb{R}^{1 \times b},
    \quad \textrm{and} \\
    & \bm{z}_{x,y} = \text{MLP} \, \, [\bm{x} \, \, \oplus \, \, \bm{y}] \in \mathbb{R}^{1 \times a},  
\end{align}
where $\oplus$ denotes the concatenation operation of the two vectors. $\bm{z}_{y,x}$ and $\bm{z}_{x,y}$ signifies the learned mutual attention distributions over the inputs $\bm{y}$ and $\bm{x}$ respectively. These attention distributions are used to take a weighted sum on the corresponding input feature vectors to generate mutually attended feature representation $\bm{y}_{m,yx}$ and $\bm{x}_{m,xy}$,
\begin{align}
\label{eq:mutual_weighted_sum}
    & \bm{y}_{m,yx} = \sum_{i=1}^b (z^i_{y,x} \, \,  \bm{y}^i) \in \mathbb{R}^{d_y},
    \quad \textrm{and} \notag \\
    & \bm{x}_{m,xy} = \sum_{i=1}^a (z^i_{x,y} \, \,  \bm{x}^i) \in \mathbb{R}^{d_x},    
\end{align}
where, $\bm{x}^i$ and $\bm{y}^i$ denote the $i^{th}$ row from the feature embeddings $\bm{x}$ and $\bm{y}$ respectively.

\textbf{Self Attention:} For the self attention component, we follow the guided self attention module used in \cite{yu2017multi}. 
The input feature $\bm{x}$ is fed to a Transformer employing multi-head attention~\cite{vaswani2017attention} to learn an attention distribution $\Psi_x$. The other input $\bm{y}$ undergoes a similar multi-head attention like $\bm{x}$, except the \emph{query} input is replaced with $\Psi_x$, which allows the model to learn $\bm{x}$-guided self attention distribution over $\bm{y}$, denoted as $\Psi_{y,x}$. $\Psi_x$ and $\Psi_{y,x}$ are passed through separate fully-connected layers for dimensionality reduction and we get $\Psi_{x} \in \mathbb{R}^{a}$ and $\Psi_{y,x} \in \mathbb{R}^{b}$. These self attention maps are used to take a weighted sum over the corresponding feature representations and generate self attended features $\bm{y}_{s,yx}$ and $\bm{x}_{s,y}$,
\begin{align}
\label{eq:self_weighted_sum}
    & \bm{y}_{s,yx} = \sum_{i=1}^b (\Psi^i_{y,x} \, \,  \bm{y}^i) \in \mathbb{R}^{d_y},
     \quad \textrm{and} \notag \\
    & \bm{x}_{s,y} = \sum_{i=1}^a (\Psi^i_{x} \, \, \bm{x}^i) \in \mathbb{R}^{d_x}.   
\end{align}

In practice, we employ two MSA modules, where we feed $\bm{q}, \, \bm{v}$ to the first one and $\bm{q}, \, \bm{r}$ to the other. The intuition behind this is the first MSA module learns to identify which region of the image, and words of the question are important to answer the question. Similarly, the second MSA module tries to identify the salient relationship features and question parts for answering the question. For both MSA blocks, we pass question features as $\bm{x}$ which guides the attention learning process of the input. This is particularly important as the question sets the objective of the task, and the quality of the learned attention distribution depends more on the question than the other inputs. Thus the first MSA module outputs $\bm{v}_{m,vq}, \bm{q}_{m,qv}, \bm{v}_{s,vq}, \bm{q}_{s,v}$ and the second one outputs $\bm{r}_{m,rq}, \bm{q}_{m,qr}, \bm{r}_{s,rq}, \bm{q}_{s,r}$. 

\subsection{Attention Fusion}
\label{sec:att_proj}
We perform multimodal attention fusion on the outputs of the MSA blocks. Each attended feature is projected to an intermediate space through fully connected layers followed by summation. As the attended features already capture rich feature description, we only use such a simple linear summation technique to capture their interaction before making the final answer prediction. The summed feature vector is then projected to the answer prediction space $d^{|\mathcal{A}|}$ through another fully connected layer where we minimize a cross-entropy loss to predict correct answer from the candidate answer set.

\section{Experiments}
We perform experiments on two large-scale VQA datasets, namely VQAv2~\cite{Goyal_2017_CVPR} and GQA~\cite{hudson2019gqa}. We train the visual relationship detector in the SRP module on VRD dataset with a VGG16 backbone, and use this pretrained model to infer relationship triplets. 

\subsection{Dataset}
We perform experiments on two large-scale VQA datasets, namely VQAv2~\cite{Goyal_2017_CVPR} and GQA~\cite{hudson2019gqa}. The VQAv2 has 200K images and 1.1M crowed-sourced questions. This is the biggest manually annotated VQA dataset. Further, GQA contains 11K images and 22M auto-generated questions, making it a more challenging evaluation setting. 

\subsection{VQA Model Architecture}
The visual feature dimension is $d_v = 2048$ for each object. To extract the semantic features from the question and relationship triplet, we use a pretrained  \texttt{\small bert-large-cased}\footnote{\url{https://huggingface.co/bert-large-cased}} model. Since a cased version is used, we do not convert the question or relationship triplets to lowercase. The extracted semantic feature dimensions for question and relationship are $d_q = d_r = 1024$. Following the recommendation in \cite{vaswani2017attention, yu2017multi}, the intermediate dimensions $d$ of the multi-head attention in transformer module is set to $512$ with $8$ heads and latent dimension of $64$. Adam optimizer~\cite{kingma2014adam} with $\beta_1 = 0.9$ and $\beta_2= 0.98$ is used. 

\sidecaptionvpos{table}{c}
\begin{SCtable*}[]
  \centering 
  \scalebox{.95}{
  \begin{tabular}{lcccccc}
    \toprule
    								& \multicolumn{6}{c}{GQA Validation Set}            \\
    \cmidrule(r){2-7}
    Methods									& Acc.$\uparrow$      & Binary$\uparrow$       & Open$\uparrow$      & Validity$\uparrow$  & Plaus.$\uparrow$    & Dist.$\downarrow$\\
    \midrule																				
    MCAN$^{\ddagger}$\cite{yu2019deep}  & 65.00     & 82.08     & 48.98     & 94.91    & 91.42      & 4.21\\
    \midrule
    $\bm{r}^{vis}+\bm{q}$                               & 51.89     & 69.02     & 35.83     & 95.13     & 91.78     & 7.34\\
    $\bm{r}^{sem}+\bm{q}$                               & 50.37     & 63.66     & 37.91     & 95.03     & 91.83     & 13.06 \\
    \midrule
    $\bm{r}^{vis}+\bm{v}+\bm{q}$                       & 58.62     & 73.25     & 44.91     & 94.95     & 91.05     & 12.63  \\
    $\bm{r}^{sem}+\bm{v}+\bm{q}$                       & 65.93     & 82.35     & 49.27     & 94.98     & 91.57     & 4.88  \\
    \midrule
    \midrule
    	                        & \multicolumn{6}{c}{Oracle Setting on GQA Validation Set} \\
    \midrule
    $\bm{r}^{oracle}+\bm{q}$                      & 68.71     & 71.84     & 68.71     & 94.94    & 92.99      & 7.29 \\
    $\bm{r}^{oracle}+\bm{v}+\bm{q}$                    & 81.15     & 85.06     & 77.48     & 95.34     & 94.26     & 1.08  \\
    \bottomrule
  \end{tabular}}
   \vspace{0.5em}
  \caption{\emph{On establishing the benefit of semantic relationship parsing for VQA.} We note that using semantic relationship features gives better performance as compared to the visual relationship features (rows 2-5). To demonstrate the richness of semantic features, we also report the upper-bound (oracle case in the last two rows), where our model delivers an absolute gain of $\sim$16 accuracy points over the MCAN \cite{yu2019deep} model. $^{\ddagger}$ For a fair comparison, the MCAN model reported here is the \texttt{MCAN-large (frcn+bbox)} version which uses bounding-box coordinates of the object proposals.
  }
  \label{tab:gqa-val-gt}
\end{SCtable*}

\subsection{Semantic vs. Visual Relationship Feature}
\label{sec:vis_vs_sem}
In Tab.~\ref{tab:gqa-val-gt}, we first establish the benefit of our proposed semantic relationship feature modeling. 
For a fair comparison with our proposed MSA model which uses bounding box coordinates of subjects and objects, we compare it with a version of MCAN \cite{yu2019deep}\footnote{official implementation of \texttt{MCAN-large (frcn+bbox)} model is at \url{https://github.com/MILVLG/openvqa}}
that also uses bounding box coordinates.
To this end, we compare the VQA performance between `semantic' and `visual' relationship features to showcase the comparative advantage on the GQA validation dataset. To develop the baseline model with visual relationship features, we train the SRP module (Sec.~\ref{sec:srp}) on the VRD dataset~\cite{lu2016visual}  with $100$ objects and $70$ predicate categories, and output visual feature of the subject and object relationship proposal alongwith relationship triplet. The visual feature of the subject and object proposals are concatenated and considered as visual relationship feature $\bm{r}^{vis}$, and the default semantic relationship features (denoted by $\bm{r}^{sem}$) are extracted from the relationship triplet. The models in Tab.~\ref{tab:gqa-val-gt} employ only the guided self-attention part of the MSA module for simplicity. 

\emph{Blind models trained with visual relationship feature perform slightly better.} In Tab.~\ref{tab:gqa-val-gt}, we see that a VQA model trained with only visual relationship features (row 2) performs better than the model trained only with semantic relationship features (row 3). This is because when the visual feature $\bm{v}$ is not available, the $\bm{r}^{sem}+\bm{q}$ model is \emph{blind} to the image and the answer prediction is based only on the relationship labels. On the other hand, the $\bm{r}^{vis}+\bm{q}$ model can \emph{see} the image as a set of the visual feature of subject-object proposals, thus performs better than the completely \emph{blind} model (row 3). However, in this extreme setting, the \emph{blind} model perform reasonably well only relying on the semantic relationship labels.

\emph{Non-blind models trained with semantic relationship feature perform significantly better.} When the visual feature is available, the VQA model with the complementary semantic relationship feature performs significantly better ($7.34\uparrow$) than its counterpart (rows 4, 5 in Tab.~\ref{tab:gqa-val-gt}). This demonstrates the complementary effectiveness of the semantic relationship features, since both these settings are identical except for the nature of the relationship feature.



\subsection{Oracle Setting}
\label{sec:oracle}
We simulate an oracle setting to further evaluate the effectiveness of using semantic relationships for VQA.  We build this setting using scene-graph annotations available for GQA \cite{hudson2019gqa} train and validation sets. Each scene-graph entry consists of ground-truth subject, 
relationship and object label. We use a scene-graph parser which converts each scene-graph entry into a list of semantic relationship triplets similar to the output of visual relationship detector of Sec.~\ref{sec:srp}, and denote the extracted semantic ground-truth relationship features as $\bm{r}^{oracle}$. 

\emph{Both blind and non-blind oracle models significantly outperform the SOTA.} 
The \emph{blind} VQA model with a ground-truth relationship label $\bm{r}^{oracle}$ achieves an overall accuracy gain of $3.71\uparrow$ compared to the state-of-the-art MCAN \cite{yu2019deep} model which is a \emph{non-blind} model (comparing rows 6 and 1 of Tab.~\ref{tab:gqa-val-gt}). This is an interesting finding showing if good enough semantic relationship label are available, the VQA model could achieve better performance than SOTA without even \emph{looking} at the image. Further, when visual feature of the image is available in the oracle setting, the model achieves $16.43 (\sim 25\%)$ accuracy gain over \cite{yu2019deep}.

\emph{`Open-ended' questions are answered better.} 
Both oracle models report significant accuracy gain ($19.73\uparrow$ and $28.5\uparrow$ compared to MCAN) for the challenging `Open' question category. These open-ended questions require diverse and broad reasoning ability to answer correctly. This is a significant finding as it sheds light upon effectiveness of using complementary semantic relationship features as an important line of research to break the bottleneck of VQA models mostly focusing on learning better visual representations.

\subsection{Ablation study}
\label{sec:ablation}
\sidecaptionvpos{table}{c}
\begin{SCtable*}[]
  \centering
  \scalebox{.95}{
  \begin{tabular}{l c cccc ccc}
    \toprule
    && \multicolumn{4}{c}{VQA-v2 Test-dev} & \multicolumn{3}{c}{GQA Test-dev}\\
    \cmidrule(r){3-6} \cmidrule(r){7-9}
    Input       & Attention     & Accuracy  & Y/N   & Number  & Other
    						    & Accuracy  & Binary    & Open\\ 
    \cmidrule(r){1-1} \cmidrule(r){2-2}	\cmidrule(r){3-6} \cmidrule(r){7-9}
                & Mutual        & 44.00 & 66.48 & 31.49 & 27.21 & 35.90 & 54.72 & 19.93 \\ 
    $\bm{r}+\bm{q}$       & Self          & 53.35 & 74.16 & 35.87 & 39.46 & 42.72 & 64.42 & 29.38\\
                & MSA           & 53.66 & 74.71 & 36.65 & 39.70 & 45.53 & 64.00 & 29.86\\
    \midrule            
                & Mutual        & 45.74 & 57.18 & 34.78 & 29.74 & 37.20 & 56.82 & 20.56\\
    $\bm{v}+\bm{q}$       & Self          & 70.14 & 86.57 & 51.59 & 60.28 & 57.03 & 76.02 & 40.76\\
                & MSA           & 70.38 & 86.78 & 52.05 & 60.59 & 57.45 & 77.08 & 40.79\\
    \midrule								
                & Mutual        & 48.29 & 67.17 & 33.24 & 35.49 & 39.24 & 55.45 & 25.48\\
    $\bm{v}+\bm{r}+\bm{q}$     & Self          & 70.46 & 87.14 & 51.26 & 60.57 & 57.72 & 76.12 & 40.48\\
                & MSA           & 70.76 & 87.10 & 53.21 & 60.77 & 58.37 & 77.70 & 40.44\\
    \bottomrule
  \end{tabular}
  }
 \vspace{.4em}
  \caption{\emph{Ablation study of our MSA model.} Ablation study of the proposed MSA model on VQAv2 Test-dev and GQA Test-dev set with different combination of input visual and semantic feature representation, and attention mechanism.}
  \label{tab:ablation_vqa_gqa}
\end{SCtable*}

We perform extensive ablation on the VQAv2 test-dev and GQA test-dev dataset{s} and report the results in Tab.~\ref{tab:ablation_vqa_gqa}. Our  goal here is to identify which input and attention combination contributes to the overall performance of our model. This is a comprehensive setup as the VQA dataset and GQA dataset consist of natural crowed-sourced and auto-generated questions respectively. We use semantic relation features for all our experiments (i.e., $\bm{r}=\bm{r}^{sem}$). 

\emph{Semantic relationship features provide accuracy boost when used in complement with visual features.}
The \emph{blind} model which only uses parsed relationship features without any visual features (rows 1--3) performs worse compared to other models that explicitly use visual features. However, when used in complement with image and question features (rows 7--9), it helps models achieve better performance on both VQA and GQA datasets.

\emph{Guided self attention provides rich attention distribution over its inputs compared to mutual attention.} For the three input combinations listed in Tab.~\ref{tab:ablation_vqa_gqa}, we ablate the MSA module by only activating mutual or self attention module. We can see that when only guided self attention module is activated, a better VQA accuracy is achieved in both the datasets. 
This is because the self attention module captures rich semantics over the the input features through its multi-head attention architecture. The mutual attention component works best in a VQA setting when the attention distribution is learned on the visual feature, which undergo{es} a second multimodal fusion with the question feature \cite{fukui2016multimodal, Farazi_2018_BMVC, benyounescadene2017mutan, ben2019block}. By design, we want the mutual attention module to capture the multimodal interaction between the inputs and feed it to the attention fusion module (Sec.~\ref{sec:att_proj}) for combining with other attention distribution{s}. Thus a standalone setup for our mutual attention module performs sub-par to the guided self attention module.

\emph{MSA module with both mutual and self attention performs best.} The full MSA model with both mutual and self attention modules achieve{s} better performance compared to when a single block is activated (rows 3,6,7 in Tab.~\ref{tab:ablation_vqa_gqa}). The mutual attention module provides complementary information that help{s} in cases where self attention alone is not sufficient. 
\subsection{Comparison with state-of-the-art models}
\label{sec:comp_with_sota}
\begin{table}[]
  \centering
  \scalebox{0.9}{
  \begin{tabular}{lcccc}
    \toprule
        								 & \multicolumn{4}{c}{VQAv2 Test-Standard}\\	
                                         \cmidrule(r){2-5}
    Methods 							 & Acc.  & Y/N   & Num.  & Other\\
    \midrule
    Ours                                 & \textbf{71.1} & 87.3 & 53.3 & 61.1\\
    \midrule
    MCAN \cite{yu2019deep}$^{\dagger}$   & 70.9 & -  & -  & -\\ 
    ReGAT \cite{Li_2019_ICCV}$^{\dagger}$& 70.6  & -  & -  & -\\ 
    Ban+Counter \cite{kim2018bilinear}$^{\dagger}$
                                         & 70.4 & -  & -  & -\\
    DFAF$^{\dagger}$\cite{gao2019dynamic}
                                         & 70.3 & - & - & -\\
    MuRel \cite{cadene2019murel}         & 68.4  & -  & -  & -\\
    Counter \cite{zhang2018vqacount}     & 68.4  & 83.6  & 51.4 & 59.1 \\
    RAF \cite{Farazi_2018_BMVC}          & 67.4  & 84.2 & 44.4 & 58.0 \\
    QAA \cite{farazi_2020_ICPR}          & 67.0  & 83.8 & 45.9 & 57.1 \\
    Graph Learner \cite{norcliffe2018learning}
                                         & 66.2  & 82.9  & 47.1 & 56.2 \\
    Bottom-Up \cite{Anderson2017up-down} & 65.7  & 82.2  & 43.9 & 56.3 \\
    \bottomrule
  \end{tabular}}
  \vspace{0.5em}
  \caption{\emph{Benchmarking our MSA model on the VQAv2 Test-Standard dataset.} Comparison of our single MSA model trained only on the VQAv2 dataset with comparable state-of-the-art models (\ie, without VL pre-training). Our approach performs favorably well against the existing VQA models. $\dagger$ models undergo additional training on the Visual Genome~\cite{krishna2016visual} dataset which provides an additional gain.}
  \label{tab:vqa-test}
\end{table}

We report the performance of our single MSA model on benchmark VQAv2 Test-Standard dataset in Tab.~\ref{tab:vqa-test}. We show that by leveraging the semantic relationship features, our model is able to outperform other comparable state-of-the-art models, even without additional training on Visual Genome dataset \cite{krishna2016visual}. Some recent models resort to ensembling with data augmentation techniques \cite{yu2019deep} or use pretrained model trained on different Vision-Language tasks and/or datasets \cite{tan2019lxmert, li2020oscar, chen_uniter_ECCV2020} to achieve superior performance. However, such approach is tangential to the motivation of this paper and are not directly comparable to our proposed model. \cite{zhang2019empirical, Hu_2019_ICCV} do not report their performance on the VQAv2 dataset and are thus not included here for comparison. We can see from Tab.~\ref{tab:vqa-test}, that our model achieves state-of-the-art accuracy on the benchmark VQAv2 dataset with comparable methods where the performance boost is achieved by incorporating semantic relationship features.

\begin{figure*}[!ht]
\begin{center}
\includegraphics[width=.68\textwidth]{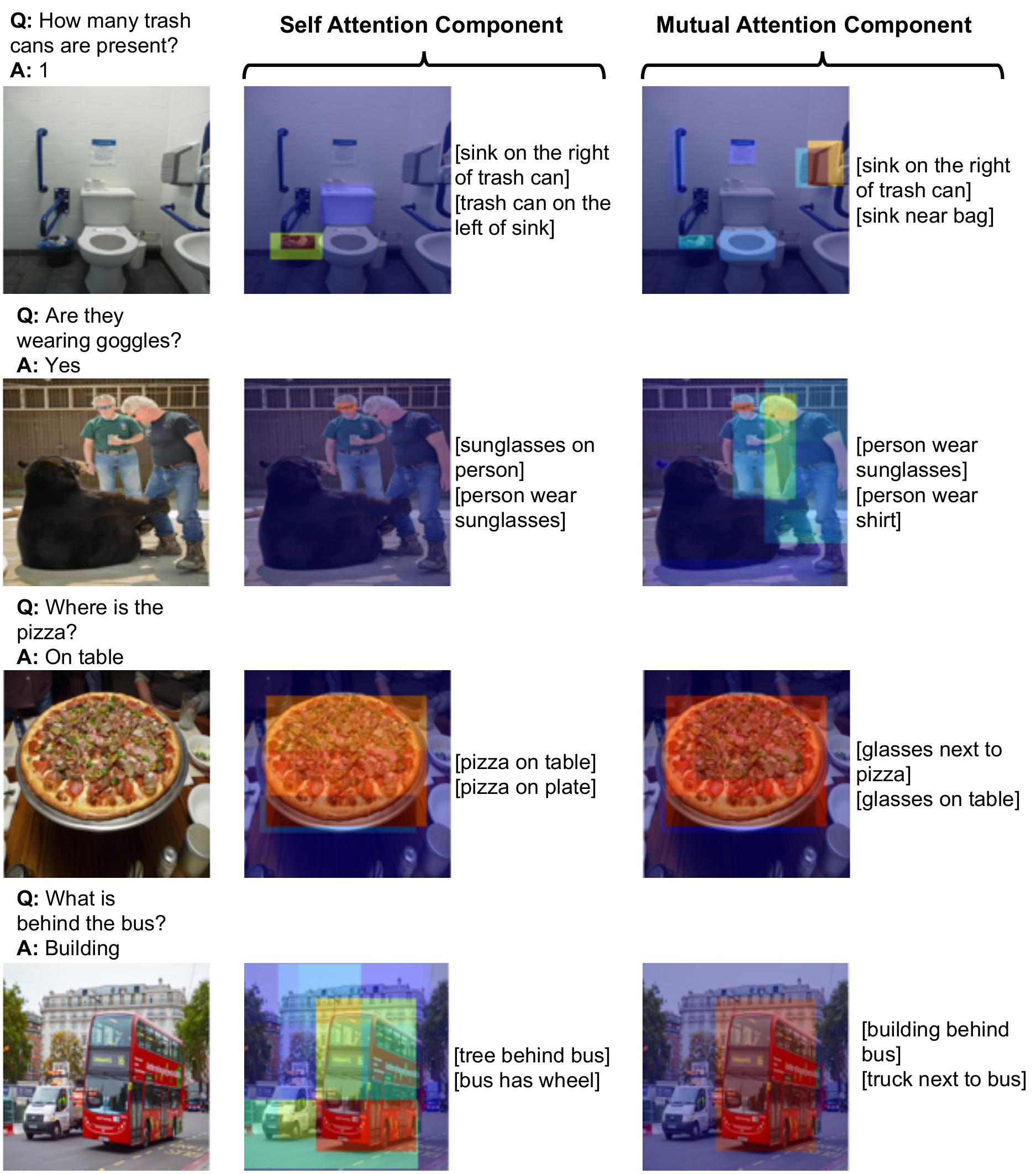}
\end{center}
\caption{\emph{Qualitative results.} Predicted answers and attention distribution over visual features ($v$) and semantic relationship features ($r$) employing our MSA model on VQAv2 dataset. We activate both Self and Mutual Attention module and separately visualize their output attention distribution over $v$ and $r$. The attention distribution over the region proposals are visualized with a heat map and two relationship triplet with highest probability is reported. The self and mutual attention component provide complementary information for predicting the correct answer.}
\label{fig:qual}
\end{figure*}

\subsection{Qualitative results}
We provide some qualitative results in Fig.~\ref{fig:qual} of MSA model on VQAv2 dataset. We visualize the attention distribution over the region proposals and list two relationship triplet with highest attention for better visualization. For simplicity we do not visualize the mutual and self attention distribution over the question words. We can see that the self and mutual attention component provide complementary attention distribution over the input features. For example, in the second row, when asked \emph{`Are they wearing goggles?'} The visual self attention component focuses more on the \texttt{sunglass} of the person on the left. The mutual attention component looks at the person on the left and the right. Similarly, the self attention component gives more attention to a relationship triplet with \texttt{sunglass} and \texttt{person}, but further looks at \texttt{person wear shirt} relationship triplet for getting more semantic context. Such complementary relationship between various attention components helps VQA model to reason better over its input feature representations.

\section{Conclusion}
VQA problem demands an in-depth understanding of the visual and semantic domains. Existing approaches generally focus on deriving more discriminative visual features or modeling the complex multi-modal interactions. Some models resort to expensive and customized pre-training on other Vision-Language tasks. In this paper, we show that an important missing piece in the existing models is that of enriched semantic relationship modeling. We demonstrate that under an oracle setting, these semantic relationships can bring the performance on par with human-level accuracy on VQA task. Further, we propose an automatic semantic relationship parser alongside a complimentary attention mechanism that delivers consistent improvements on SOTA across two challenging VQA datasets. Our results strongly advocate for further investigation on better relationship modeling in the semantic domain, a direction less explored so far in the VQA community. 

{\small
\bibliographystyle{ieee_fullname}
\bibliography{egbib}
}

\end{document}